\pdfoutput=1

\documentclass[11pt]{article}

\usepackage{coling}

\usepackage[T1]{fontenc}

\usepackage[utf8]{inputenc}

\usepackage{inconsolata}

\usepackage{times}
\usepackage{latexsym}
\usepackage{url}
\usepackage{amsmath} 

\usepackage{pifont}
\usepackage{dingbat}
\usepackage{amssymb}

\usepackage{graphicx}

\usepackage{array}
\usepackage{makecell}
\usepackage{colortbl}
\usepackage{boldline}
\usepackage{multirow}
\usepackage{xcolor}

\usepackage{tikz}
\usetikzlibrary{calc} 

\usepackage{multicol}

\usepackage{mathtools} 
\usepackage{hyperref}

\usepackage{microtype}

\usepackage{enumitem}
\setlist[itemize,1]{left=0.2em, label=$\bullet$, itemsep=1pt}

\usepackage{relsize} 
\renewcommand{\footnotesize}{\scriptsize} 

\usepackage{xspace}

\usepackage{graphicx}
\usepackage{caption}
\usepackage{subcaption}
\usepackage{float} 

\usepackage{array}
\newcolumntype{M}[1]{>{\centering\arraybackslash}m{#1}}

\usepackage{tabularx}

\title{Embedding Style Beyond Topics:\\ Analyzing Dispersion Effects Across Different Language Models}

\author{{\bf Benjamin Icard$^{1}$,} Evangelia Zve$^{1,2}$,  {\bf Lila Sainero$^{1}$,} \\ {\bf Alice Breton$^{1}$,} {\bf and Jean-Gabriel Ganascia$^{1}$} \vspace{0.1in}\\ 
$^{1}$ LIP6, Sorbonne University, CNRS, France \\
$^{2}$ Infopro Digital, France}

\begin{document}
\maketitle

\begin{abstract}

This paper analyzes how writing style affects the dispersion of embedding vectors across multiple, state-of-the-art language models. While early transformer models primarily aligned with topic modeling, this study examines the role of writing style in shaping embedding spaces. Using a literary corpus that alternates between topics and styles, we compare the sensitivity of language models across French and English. By analyzing the particular impact of style on embedding dispersion, we aim to better understand how language models process stylistic information, contributing to their overall interpretability.

\end{abstract}

\section{Introduction}

In recent years, large language models (LLMs) have shown advanced natural language processing capabilities across diverse tasks, making their explainability an important area of research \cite{zhao2024explainability}. A key aspect of these models is their ability to generate meaningful text representations through \textit{vector embeddings}, that encode semantic information. Although topic modeling along embedding representations has been widely studied \cite{peinelt2020tbert}, the influence of writing style on these representations has received less attention \cite{terreau2021writing, chen2023writing}. By leveraging sophisticated neural architectures \cite{achiam2023gpt, jiang2023mistral}, current large-scale models, such as those developed by OpenAI and Mistral, provide new investigative paths in that respect. 

This paper aims to provide deeper insights into \textcolor{black}{how different language models encode writing style and study their sensitivity to stylistic features.} Specifically, we seek to examine the relative impact of style versus topic on the spatial dispersion of embedding vectors, across different language models in both French and English. Our primary focus is on the particular influence of writing style, with an emphasis on explainability. 

To conduct this analysis, we designed an experimental study that systematically interchanged topic and style dimensions using text generation techniques. We selected two established literary works as raw material: Raymond Queneau's \textit{Exercices de Style} \cite{queneau1947exercices} and Félix Fénéon’s \textit{Nouvelles en trois lignes} \cite{feneon1970oeuvres}. \textit{Exercices de Style} is a highly original piece of experimental literature in which Queneau writes numerous stylistic variations of a single narrative (a brief confrontation between bus passengers) while maintaining the same topic across all versions. By contrast, \textit{Nouvelles en trois lignes} covers a wide range of topics (e.g., political events, crime, nature) while keeping a consistent style marked by a vivid and ironic tone. To enrich this material, we employed text generation techniques. We created a corpus where Queneau's style aligns with Fénéon’s unique style, and another corpus where Fénéon’s style varies in line with Queneau’s plurality of styles. This design aimed to effectively assess the impact of topic and style on embedding dispersion. 

Section \ref{sec:related} reviews existing work on the computational analysis of writing style, contrasting it with topic modeling, with an emphasis on vector embedding techniques. Section \ref{sec:corpus} describes the \textsc{Queneau-Feneon} dataset, a collection of textual documents compiled for topic and style experimentation, employing a specific text generation methodology. Section \ref{sec:experiments} describes the experimental tasks and results obtained on this dataset, including clustering to assess alignment with predefined classes, analysis of how style and topic influence embedding dispersion, and the identification of key linguistic features that may explain this dispersion. Finally, Section \ref{sec:conclusion} concludes our investigation and outlines directions for future work.


\section{Related Work}
\label{sec:related}

Embedding vector representations have been primarily studied in the context of topic modeling, building on BERT studies \citep{devlin2018bert}. Techniques using word embeddings or sentence vectors, such as SBERT \citep{reimers2019sentence}, were developed to extract topics from textual documents, outperforming traditional statistical topic modeling methods like Latent Dirichlet Allocation (LDA) \citep{blei2003latent}. A notable advancement in that respect is BERTopic, which refines these different methods \citep{grootendorst2022bertopic}. Recent progress has focused on combining traditional methods like LDA with word embeddings, resulting in improved topic quality metrics and interpretability \citep{dieng2020topic}.

\textcolor{black}{Currently, research on embedding vector representations of writing style remains relatively underexplored compared to topic modeling \citep{dai2019style}. Existing computational and statistical approaches to writing style \citep{herrmann2021computational}, including stylometry, focus on features such as word frequency, part-of-speech tags, N-grams \cite{rios2022detection}, specific lexical entries and punctuation \citep{faye2024exposing,icard2024multi}, TF-IDF \citep{bui2011writer}, and vector embeddings \cite{chen2023writing}. From a literary perspective, these approaches consider writing style as a manifestation of an author's unique voice and aesthetic choices \citep[e.g.][]{verma2019lexical,mani2022computational}.}

In continuation of computational stylometry, and enabled by transformer architecture \citep{hao2021self}, recent studies have examined stylistic features using embedding techniques \citep{liu2024team}, with particular focus on literary texts \citep{maharjan-etal-2019-jointly}, but also in other domains, like news media and Generative AI \citep{bevendorff2024overview}. However, a comprehensive approach that fully captures the entire spectrum of writing style remains underdeveloped. \citet{terreau2021writing} proposed a novel evaluation framework for author verification embedding methods based on writing style, quantifying whether the embedding space effectively captures a set of stylistic features as the best proxy of an author's writing style. In addition to enhancing explainability, their work revealed that recent models are mostly driven by the inner semantics of authors' production and are outperformed by simple baselines on several linguistic axes. \citet{chen2023writing} proposed a writing style embedding method based on contrastive learning for multi-author writing style analysis, achieving promising results in detecting style changes in multi-author documents. Addressing the challenge of content-independent style representations, \citet{wegmann2022author} introduced a variation of the authorship verification training task that controls for content using conversation or domain labels, finding that representations trained by controlling for conversation are better at representing style independent from content.

While a handful of studies have explored style embedding representations from a comprehensive perspective, most existing research has primarily focused on analyzing how specific models encode targeted stylistic features, such as syntactic and lexical embeddings. This paper proposes \textcolor{black}{a structured methodology to compare} how style versus topic variation influences the embedding dispersion of various language models in both French and English.


\section{Dataset}
\label{sec:corpus}

To conduct this study, we compiled a corpus named \textsc{Queneau-Feneon}, consisting of 584 textual documents, with 292 texts in French and 292 texts in English. The corpus was developed in two stages: first, we created a \textit{reference corpus} using extant literary works by Raymond Queneau and Félix Fénéon; second, we created a \textit{generated corpus} by tranforming these original texts. The generated corpus was created using $\texttt{GPT-4o}$,\footnote{\url{https://platform.openai.com/docs/models/gpt-4o}} with a prompting methodology described below.

\subsection{Reference corpus}

We began by compiling a \textit{reference corpus} of 146 texts in each language by uniting two symmetric classes with respect to topic and style variation. The first class, named \textsc{Queneau\_ref}, contains 73 texts extracted from Raymond Queneau's \textit{Exercices de style}. These texts are written in 73 different styles but all deal with the same topic of a bus journey. The second class, named \textsc{Feneon\_ref}, also contains 73 texts, this time extracted from Félix Fénéon's \textit{Nouvelles en trois lignes}. The specificity here is that these texts include numerous topics but all share the consistent style of Feneon.

We used the original French versions of the Queneau and Fénéon's corpus~\cite{queneau1947exercices,feneon1970oeuvres} to form the French \textsc{Queneau\_ref} and \textsc{Feneon\_ref} classes. For English, we used the extant English translations of both author's corpus~\cite{queneau2013exercises,feneon2007novels} to form the English \textsc{Queneau\_ref} and \textsc{Feneon\_ref} classes. Each of these classes contains exactly 73 texts. 

Note that Raymond Queneau's \textit{Exercices de style} originally contained 99 texts; however, we retained only 73 texts to ensure a balanced comparison with Félix Fénéon's \textit{Nouvelles en trois lignes}, creating not only equally sized classes but also ensuring comparable text lengths between classes.



\subsection{Generated corpus}

We obtained a \textit{generated corpus} by applying $\texttt{GPT-4o}$ text generation on $\textsc{Queneau\_ref}$ and $\textsc{Feneon\_ref}$, respectively, in both French and English. We began by generating a class named $\textsc{Queneau\_gen}$, prompting $\texttt{GPT-4o}$ to rewrite all 73 stories of $\textsc{Queneau\_ref}$ in the uniform style of $\textsc{Feneon\_ref}$. Then, we generated a complementary class named $\textsc{Feneon\_gen}$ by prompting $\texttt{GPT-4o}$ to rewrite each of the 73 stories of $\textsc{Feneon\_ref}$ into one of the 73 different writing styles of $\textsc{Queneau\_ref}$. 
As in the reference class, the generated class contains exactly 146 texts equally divided into 73 texts for $\textsc{Queneau\_gen}$ and 73 texts for $\textsc{Feneon\_gen}$. Table \ref{tab:overview} gives a general overview of the $\textsc{Queneau-Feneon}$ corpus, obtained with the French and English prompts given in Figure \ref{fig:prompts}. 

\vspace{0.2cm}
\begin{table}[h]
\centering
\begin{scriptsize}
\renewcommand{\arraystretch}{1.23}  
\setlength{\tabcolsep}{3pt}  
\begin{tabular}{|c|c|}
\hline
\multicolumn{2}{|c|}{\textbf{Reference Corpus}}  \\ 

\multicolumn{2}{|c|}{146 texts per language}  \\
\hline

\textsc{Queneau\_ref} & \textsc{Feneon\_ref} \\ 

\textit{same topic, various styles} & \textit{various topics, same style} \\

73 texts per language & 73 texts per language\\ \hline

\textsc{Queneau\_gen} & \textsc{Feneon\_gen} \\ 

\textit{same topic, same style} & \textit{various topics, various styles}\\ 

73 texts per language & 73 texts per language\\ \hline

\multicolumn{2}{|c|}{\textbf{Generated Corpus}} \\

\multicolumn{2}{|c|}{146 texts per language}  \\

\hline
\end{tabular}

\begin{tikzpicture}[overlay, remember picture]

    \coordinate (start) at ($(3.2, 2.3) $);  %
    \coordinate (end) at ($(3.2, 1)$);    %
    \draw[->, thick, bend left] (start) to[out=180,in=180] (end)
    node[midway, left, xshift=-93pt, yshift=48pt] {\scriptsize\rotatebox[origin=c]{90}{ \textsc{Feneon\_ref}}};  

    \coordinate (start) at ($(-3.2, 2.3) $);  %
    \coordinate (end) at ($(-3.2, 1)$);    %
    \draw[->, thick, bend left] (start) to[out=180,in=180] (end) 
    node[midway, left, xshift=103pt, yshift=48pt] {\scriptsize \rotatebox[origin=c]{-90}{\textsc{Queneau\_ref}}};  %

\end{tikzpicture}
\end{scriptsize}
\caption{Overview of the \textsc{Queneau-Feneon} corpus involving text generation with $\texttt{GPT-4o}$.}\label{tab:overview}
\end{table}

\begin{figure}[h]
\begin{tiny}
\fbox{
\parbox{7.35cm}{\vspace{-0.8em}
\setlength{\leftmargini}{0.0em} 
\noindent
\setlength{\itemsep}{0pt}  
\begin{itemize}[leftmargin=0.0em]

\item[] $\textsc{Queneau\_gen}$:
\begin{itemize}
\item[—] French version: \texttt{"Ré écris ce texte : \textbackslash n" + exercice.read() + "\textbackslash n En copiant le style de Fénéon dans les 'nouvelles en trois lignes'"}

\item[—] English version: \texttt{"Re write this text in strictly less than 30 words and using only 1 to 3 sent\-en\-ces: \textbackslash n "+exercice.read() +"\textbackslash n Copying Feneon's style in 'novels in three lines'"};

\end{itemize}

\item[] $\textsc{Feneon\_gen}$: 
\begin{itemize}
\item[—] French version: \texttt{"Ré écris ce texte :\textbackslash n" + nouvelle.read() +"\textbackslash n En copiant le style de ce deuxième texte :" + exercice.read()};

\item[—] English version: \texttt{"Re write this text: \textbackslash n" + nouvelle.read() +"\textbackslash n Copying the style of this second text: "+exercice.read()}
\end{itemize}
\end{itemize}
\vspace{-0.8em}}
}
\end{tiny}
\caption{French and English $\texttt{GPT-4o}$ prompts used for generating $\textsc{Queneau\_gen}$ and $\textsc{Feneon\_gen}$, based on $\textsc{Queneau\_ref}$ and $\textsc{Feneon\_ref}$.}
\label{fig:prompts}
\end{figure}

\section{Experiments}
\label{sec:experiments}

\subsection{Corpus validation}

We performed \textit{k-means} clustering on the embeddings of the French and English $\textsc{Queneau-Feneon}$ corpus. Twelve embedding models were selected in that respect, based on the following criteria: diversity, representativeness, dimensionality, multilingual capability, computational efficiency, explainability, and high performance according to the Massive Text Embedding Benchmark (MTEB) \cite{muennighoff2022mteb} \textcolor{black}{at time of the paper submission (September 16, 2024).}\footnote{\url{https://huggingface.co/spaces/mteb/leaderboard}} The full list of tested models is presented in Table \ref{tab:meanpca} (see section of supplementary materials indicating dimensionality per model and URLs).


The clustering task aimed to assess the methodology used to build the \textsc{Queneau-Feneon} dataset. Here \textit{k-means} measures the ability of the embedding models to effectively capture, and distinguish, the different topics and styles of our corpus. Notice that our goal was not to identify the optimal number of clusters for this dataset but, this number $k$ being set to 4, to determine whether the texts in the four clusters align with the four classes of the \textsc{Queneau-Feneon} corpus. 

\begin{figure*}[t]
    \centering
    \begin{subfigure}[b]{0.47\linewidth}
        \centering
        \includegraphics[width=0.51\linewidth]{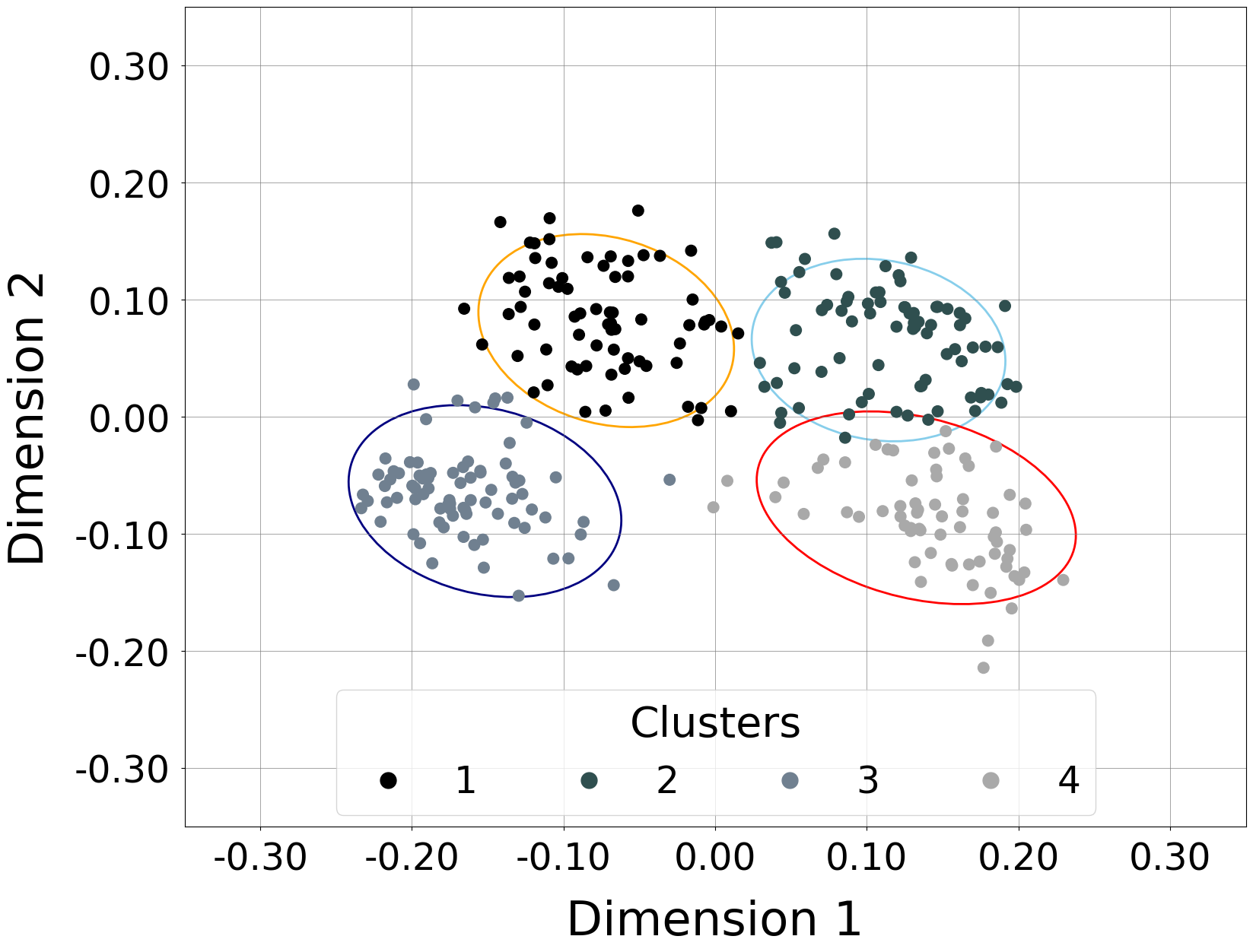}
        \hfill
        \includegraphics[width=0.47\linewidth]{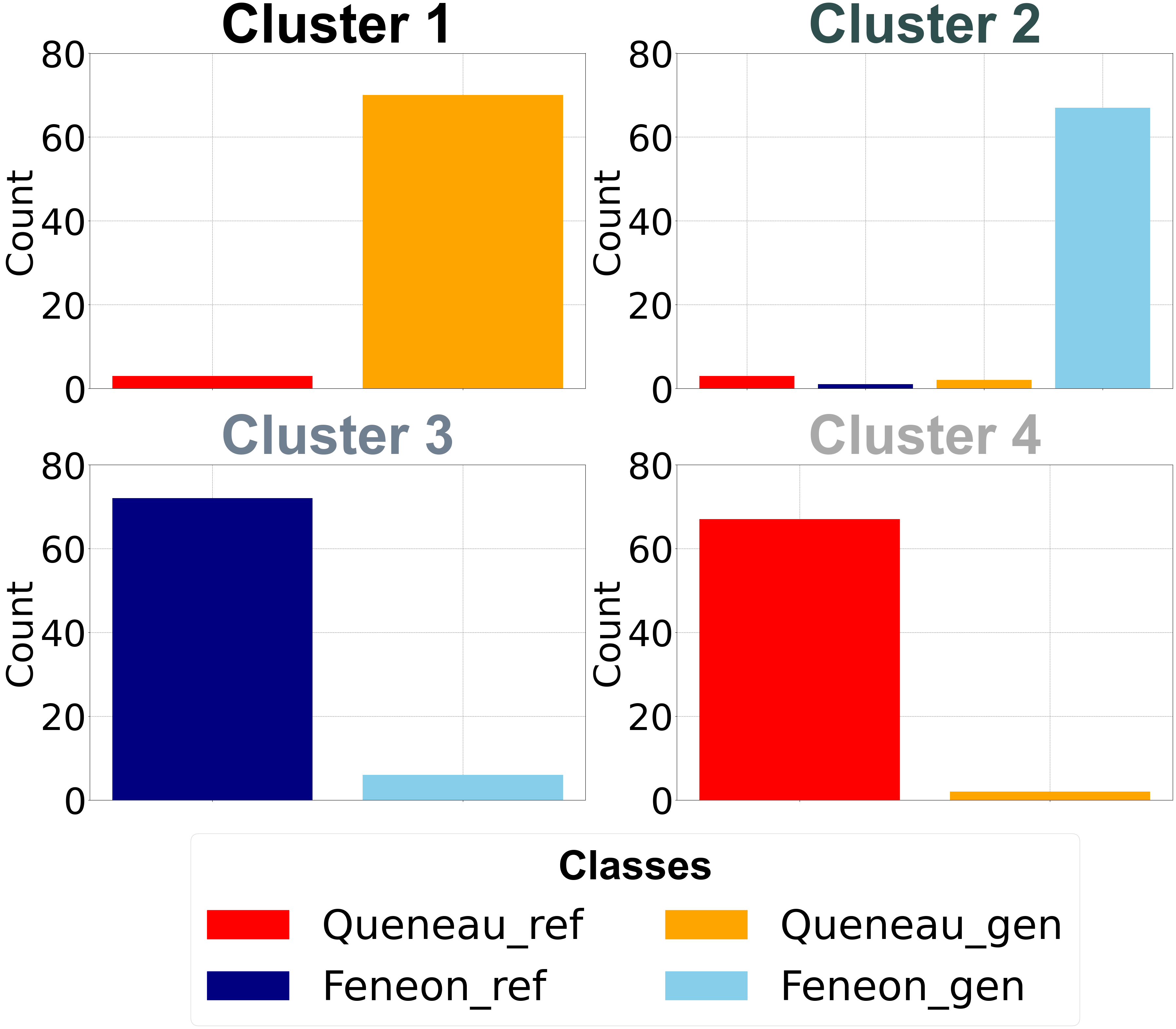}
        \caption{Projection for French \textsc{Queneau-Feneon} (left) with indication of majority class for each cluster (right).}
    \end{subfigure}
    \hspace{0.01\linewidth} 
    \rule{0.01pt}{4.3cm}
    \hspace{0.01\linewidth} 
    \begin{subfigure}[b]{0.47\linewidth}
        \centering
        \includegraphics[width=0.5\linewidth]{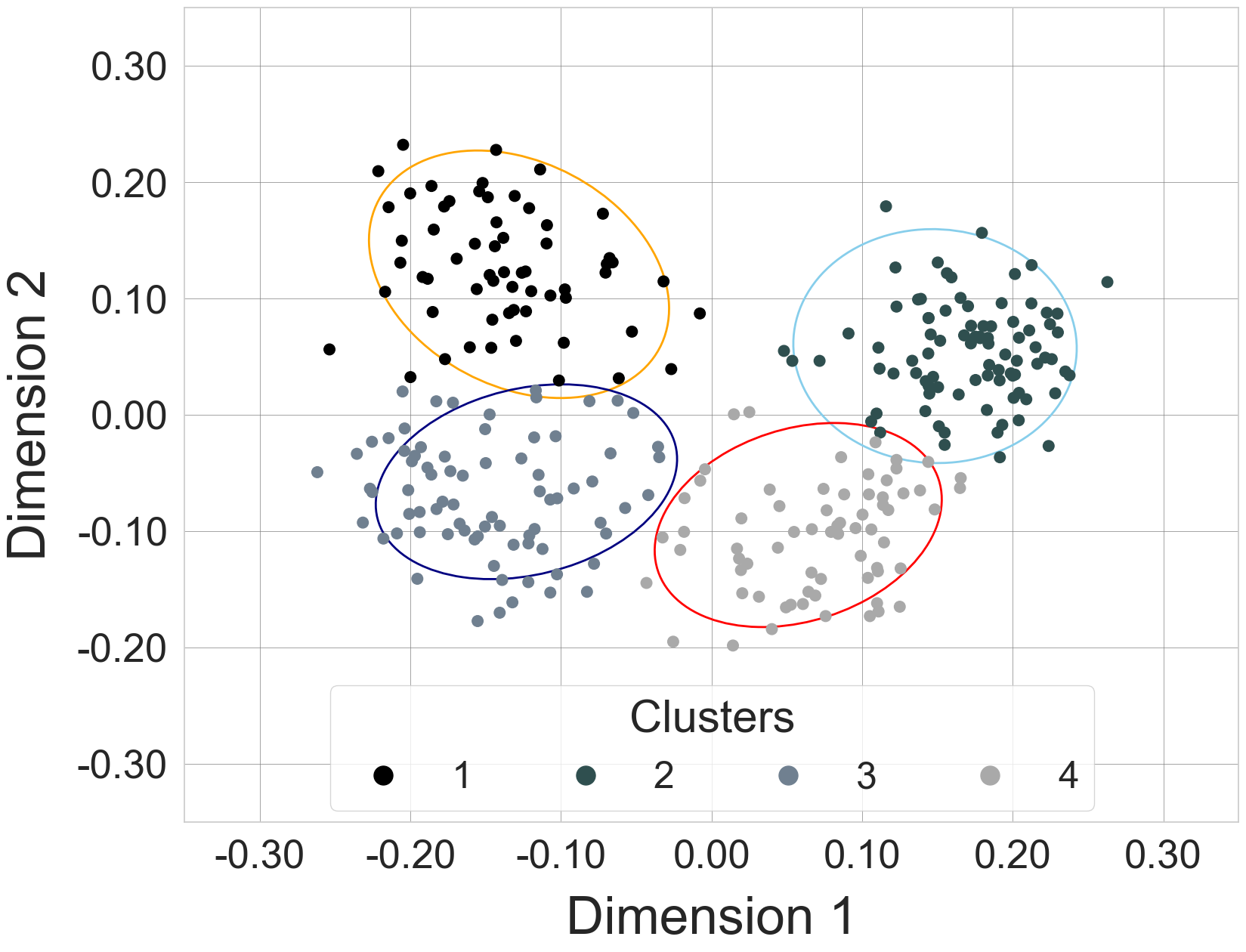}
        \hfill
        \includegraphics[width=0.45\linewidth]{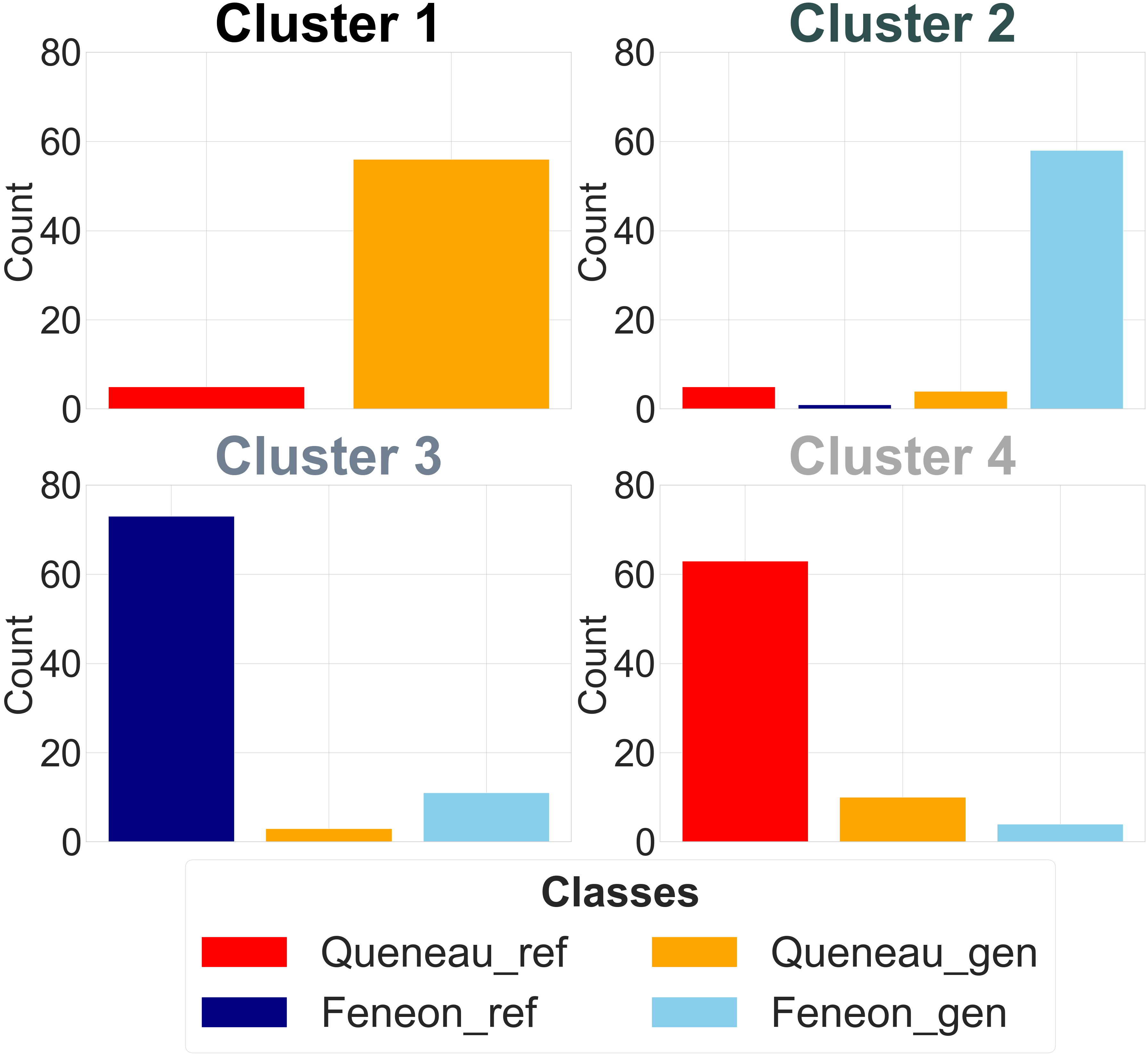}
        \caption{Projection for English \textsc{Queneau-Feneon} (left) with indication of majority class for each cluster (right).}
    \end{subfigure}
    \caption{2D PCA projection of the 4 clusters obtained with $\texttt{mistral-embed}$ on the \textsc{Queneau-Feneon} corpus and distribution of texts per cluster, for French (a) and for English (b).}
    \label{fig:mistral_clustering}
\end{figure*}







We combined two popular evaluation metrics to assess the clustering task: \textit{Purity} \cite{manning2008introduction} and \textit{NMI} (Normalized Mutual Information) \cite{danon2005comparing}. Ranging from 0 to 1, \textit{Purity} and \textit{NMI} are external cluster evaluation metrics based on the a priori knowledge of our dataset \citep{soni2024clutching}. To facilitate model comparisons across languages and dimensions, we define a qualitative score $\bar{S}^\text{D}(m)$ which, for a given model $m$, given a specific dimensionality D, averages the \textit{Purity} and \textit{NMI} scores of the model $m$ for D:  

\vspace{-0.4cm}
\begin{equation}
\label{eq:global}
\bar{S}^\text{D}(m) = \frac{{\it Purity^\text{D}(m)} + {\it NMI^\text{D}}(m)}{2}
\end{equation}

\vspace{0.15cm}




We first applied clustering on the full-dimensional embeddings obtained with the twelve selected models on the French and English \textsc{Queneau-Feneon} corpus. Subsequently, we employed Principal Component Analysis (PCA) \cite{wold1987principal} to determine the most effective dimensionality for an aligned comparison of performances across models and languages. PCA mitigates the so-called \textit{``curse of dimensionality''} \cite{Shlens2014,Jolliffe2016} that may arise with high-dimensional text embeddings, while preserving well the global structure of the vector space. Also, various studies have shown that \textit{k-means} clustering performance is notably improved by PCA \citep[e.g.,][]{Holland2020, Aliakbar2022}.

To begin with, we calculated the $\bar{S}^\text{D}(m)$ score of each model $m$ for each of the PCA dimensionality D considered: 2D, 3D, 5D, and 10D. Then, we calculated the mean $\bar{S}^\text{D}$ score of the 12 models for each of those dimensionalities to identify the best PCA combining reduction with information retention. Both languages considered, the ranking order obtained from best to worst is: 2D PCA, 3D PCA, 10D PCA, 5D PCA, followed by the FullD specific to each model. This ranking approach ensures an optimal balance between data compression and preservation of relevant information in the embedding vector space. 
Table \ref{tab:meanpca} provides the mean and median $\bar{S}^\text{D}$ scores for the top 3 best-reduced dimensionalities using PCA (2D, 3D, 10D) on French and English \textsc{Queneau-Feneon}, compared to FullD (now specific to each model).\footnote{\textcolor{black}{All the individual \textit{Purity} and \textit{NMI} scores of each model are detailed for each specific dimensionality in the GitHub of supplementary materials.}}

\renewcommand{\arraystretch}{1.15}
\begin{table}[h]
\centering
\tiny
\resizebox{\linewidth}{!}{%
\begin{tabular}{|l|c|c|c|c|}
\hline
\multicolumn{5}{|l|}{\textbf{FRENCH}} \\
\hline
 & \textbf{2D PCA} & \textbf{3D PCA} & \textbf{10D PCA} & \textbf{FullD} \\
\hline
\texttt{mistral-embed} & 0.8522 & 0.8579 & 0.6309 & 0.6713 \\
\hline
\texttt{solon-...-large-0.1} & 0.8454 & 0.8630 & 0.7067 & 0.6312 \\
\hline
\texttt{multilingual-e5-large} & 0.8285 & 0.7935 & 0.6204 & 0.6202 \\
\hline
\texttt{e5-base-v2} & 0.7406 & 0.7326 & 0.7179 & 0.6510 \\
\hline
\texttt{voyage-2} & 0.6996 & 0.6912 & 0.5766 & 0.5919 \\
\hline
\texttt{xlm-roberta-large} & 0.6666 & 0.6485 & 0.7704 & 0.6500 \\
\hline
\texttt{sentence-camembert-base} & 0.6194 & 0.5476 & 0.6639 & 0.5147 \\
\hline
\texttt{all-roberta-large-v1} & 0.5960 & 0.5189 & 0.6074 & 0.6363 \\
\hline
\texttt{distilbert-base-uncased} & 0.5632 & 0.5622 & 0.5668 & 0.5679 \\
\hline
\texttt{text-embedding-3-small} & 0.5335 & 0.5578 & 0.5005 & 0.5087 \\
\hline
\texttt{-multi...-mpnet-base-v2} & 0.5146 & 0.4575 & 0.4724 & 0.4691 \\
\hline
\texttt{all-MiniLM-L12-v2} & 0.3915 & 0.4222 & 0.5178 & 0.4117 \\
\hline
\textbf{Mean} & \textbf{0.6623} & \textbf{0.6117} & \textbf{0.6748} & \textbf{0.6633} \\
\hline
\textbf{Median} & \textbf{0.6354} & \textbf{0.5744} & \textbf{0.6917} & \textbf{0.6597} \\
\hline

\multicolumn{5}{|l|}{\textbf{ENGLISH}} \\
\hline
\texttt{solon-...-large-0.1} & 0.7779 & 0.5654 & 0.5350 & 0.6625 \\
\hline
\texttt{mistral-embed} & 0.7491 & 0.5497 & 0.5682 & 0.6547 \\
\hline
\texttt{multilingual-e5-large} & 0.6887 & 0.5696 & 0.5915 & 0.5959 \\
\hline
\texttt{voyage-2} & 0.6636 & 0.7045 & 0.5772 & 0.7519 \\
\hline
\texttt{text-embedding-3-small} & 0.6447 & 0.6551 & 0.5111 & 0.4577 \\
\hline
\texttt{all-roberta-large-v1} & 0.6429 & 0.5272 & 0.5270 & 0.5203 \\
\hline
\texttt{distilbert-base-uncased} & 0.6291 & 0.6712 & 0.6780 & 0.5335 \\
\hline
\texttt{all-MiniLM-L12-v2} & 0.5976 & 0.4966 & 0.4726 & 0.5355 \\
\hline
\texttt{e5-base-v2} & 0.5464 & 0.5280 & 0.5215 & 0.5335 \\
\hline
\texttt{-multi...-mpnet-base-v2} & 0.5211 & 0.4716 & 0.4697 & 0.4348 \\
\hline
\texttt{sentence-camembert-base} & 0.5132 & 0.5661 & 0.5312 & 0.5375 \\
\hline
\texttt{xlm-roberta-large} & 0.4693 & 0.6918 & 0.7027 & 0.6094 \\
\hline
\textbf{Mean} & \textbf{0.6260} & \textbf{0.5977} & \textbf{0.5995} & \textbf{0.6168} \\
\hline
\textbf{Median} & \textbf{0.6289} & \textbf{0.5646} & \textbf{0.5605} & \textbf{0.6119} \\
\hline
\end{tabular}}
\caption{Details of $\bar{S}^\text{D}$ scores per model for the three PCA reduction methods receiving the best mean $\bar{S}^\text{D}$ scores in both languages, compared to FullD. Models are ordered based on their 2D PCA scores.}
\label{tab:meanpca}
\vspace{-0.11cm}
\end{table}

Mean $\bar{S}^\text{D}$ scores obtained for 2D, 3D and 10D PCA are consistent, with all mean scores equal or higher than .6. For each dimension, the closeness of medians and means indicate that $\bar{S}^\text{D}(m)$ scores per model $m$ are balanced, with low or no skewness. Validation results turned out to be slightly better for French but consistent across languages and dimensions. For both French and English, the best dimension overall was 2D PCA with $\texttt{mistral-embed}$ and $\texttt{solon-embeddings-large-0.1}$ obtaining the best performances, then followed by $\texttt{multilingual-e5-large}$ and closely by $\texttt{voyage-2}$ with one rank difference. Figure \ref{fig:mistral_clustering} presents the 2D PCA projections of the clustering obtained with $\texttt{mistral-embed}$ and the correspondence with the four known classes of the \textsc{Queneau-Feneon} corpus.

Projections in 2D PCA of the embeddings obtained with $\texttt{mistral-embed}$ reveal four dense clusters in French and English, with clear separation of the four groups delineated. Distributions of texts per cluster also indicated in Figure \ref{fig:mistral_clustering}  shows the existence of a majority class matching fairly well with exactly one of the four initial classes of the \textsc{Queneau-Feneon} corpus, in line with the $\bar{S}^\text{2D}$ scores reported in Table \ref{tab:meanpca} concerning $\texttt{mistral-embed}$ (for French: $0.8522$, for English: $0.7491$). It can be observed in Table \ref{tab:meanpca} that the good correspondence obtained with $\texttt{mistral-embed}$ also transfers to other models that show high $\bar{S}^\text{D}$ scores (greater than $.65$) in both languages, such as e.g. $\texttt{solon-embeddings-large-0.1}$, 
\texttt{multilingual-e5-large}, and \texttt{voyage-2}. Support by other models is more moderate to low, with inconsistency across languages for e.g. $\texttt{e5-base-v2}$ and $\texttt{xml-roberta-large}$.








\subsection{Dispersion within classes}
\label{ssec:dispersionclass}

To gain a deeper understanding of how topic and style impact embedding representations, we analyzed embeddings dispersion within the $\textsc{Queneau-Feneon}$ corpus. Specifically, we aimed to determine whether variations in topic and in style lead to increase or decrease dispersion. Additionally, we aimed to evaluate the relative contribution of style versus topic to this effect.


To conduct this analysis, we employed the Uniform Manifold Approximation and Projection (UMAP)~\cite{mcinnes2018umap} technique for dimensionality reduction of our embedding vector space. UMAP was chosen over other methods, like PCA previously used for clustering, due to its superior ability to preserve both local and global structures within the embedding space. Additionally, t-distributed Stochastic Neighbor Embedding (t-SNE) is effective at preserving local structures, but it often distorts the global structure~\cite{anowar2021conceptual}, making it less suitable for our specific analysis. Since we aimed to ensure that local information was primarily preserved while also maintaining an accurate global structure, UMAP was the most appropriate choice, as particularly well-suited for distance-based analysis of high-dimensional text embedding spaces~\cite{cox2021directed}.

To account for the non-deterministic nature of UMAP~\cite{mcinnes2018umap}, we performed 30 applications of the model (iterations) with different random seeds, for dimensionality reductions of 2D, 3D, 5D, and 10D. The adoption of different random seeds ensures that the results are stable and not sensitive to specific initial conditions, providing a more robust estimate of embedding dispersion.

\begin{figure*}[h]
    \centering
    \begin{minipage}[t]{0.43\linewidth} 
        \centering
        \includegraphics[width=\linewidth]{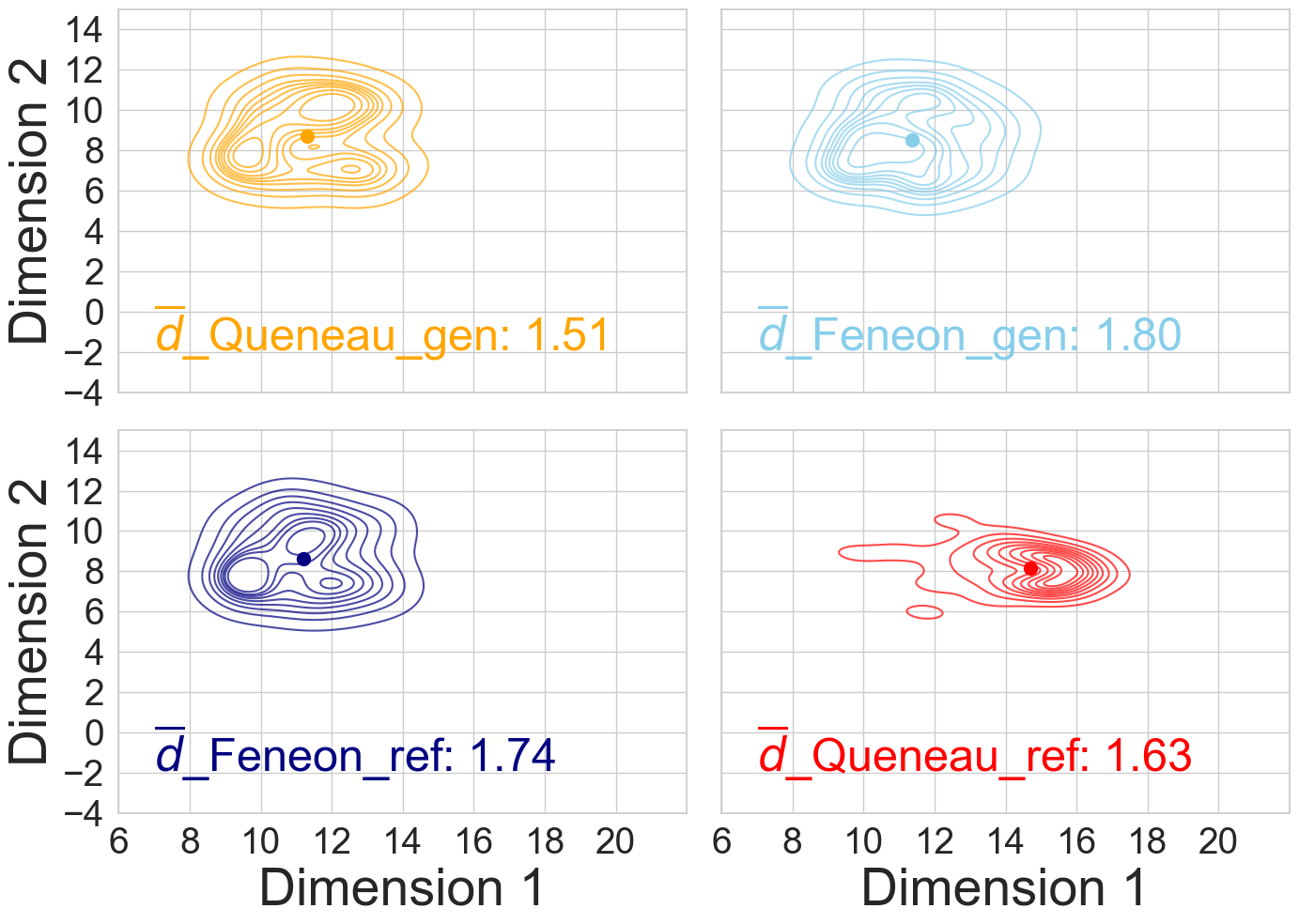}
    \end{minipage}
    \hspace{0.015\linewidth} 
    \rule{0.5pt}{5.1cm} 
    \hspace{0.015\linewidth} 
    \begin{minipage}[t]{0.43\linewidth} 
        \centering
        \includegraphics[width=\linewidth]{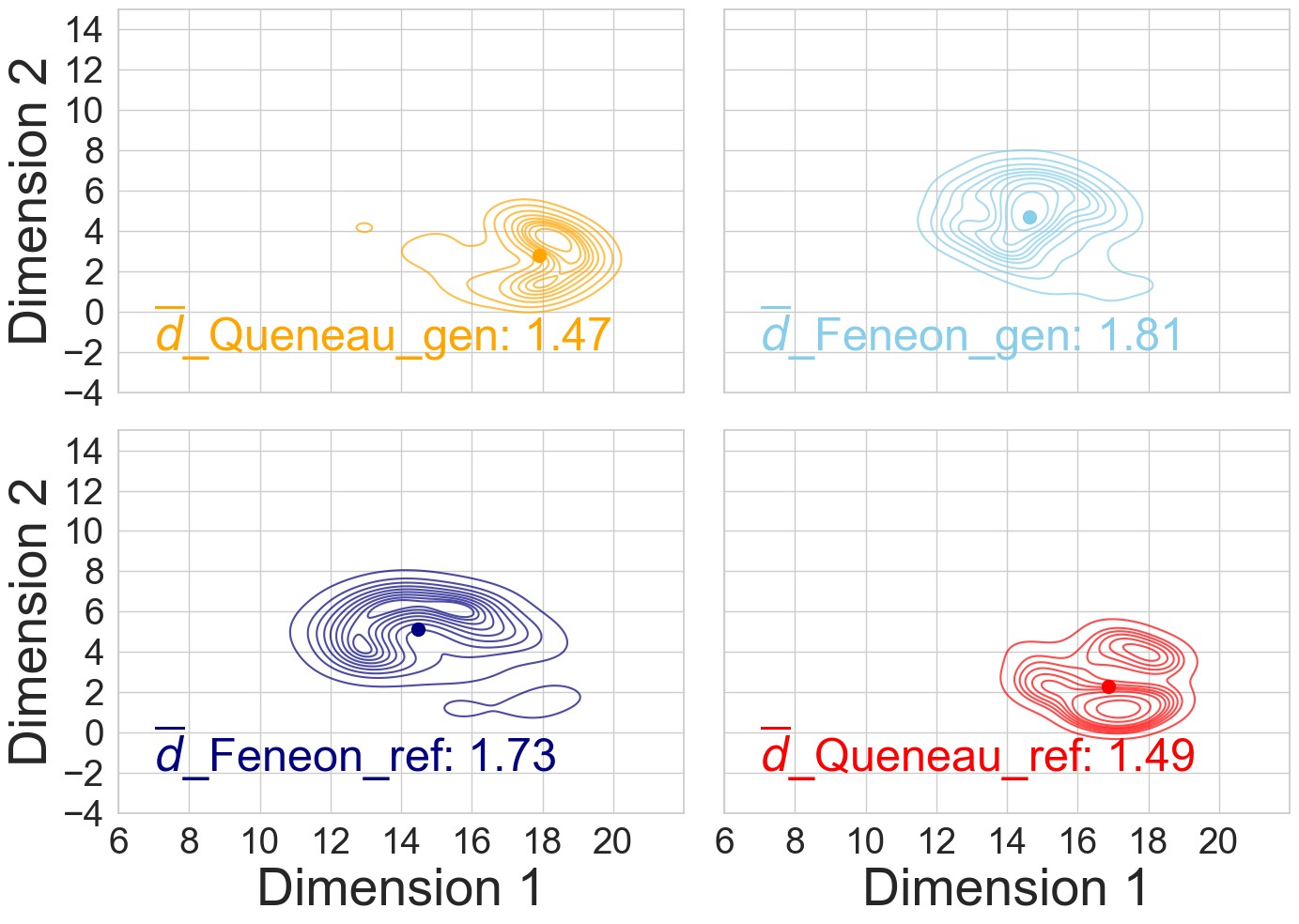}       
    \end{minipage}
\caption{\textcolor{black}{2D UMAP contour plots of the embedding dispersion obtained on the $\textsc{Queneau-Feneon}$ corpus with model $\texttt{all-MiniLM-L12-v2}$, for French (left) and for English (right). In each subplot, the overall spread of the embeddings around centroid (for the last seed) is represented by the external contour line, the isolines represent differences in densities of embedding vectors, the centroid is indicated by a dot, and $\bar{d}_X$ corresponds to the mean centroid distance for the targeted corpus $X$.}}\label{fig:dispersionplot}
\end{figure*}

\renewcommand{\arraystretch}{1.15}
\begin{table*}[h]
    \centering
    \tiny 
\resizebox{\textwidth}{!}{
\begin{tabular}{|l|c|c|c|c|c||c|c|c|c|c|}
        \hline
        \multirow{3}{*}{\textbf{DISPERSION HYPOTHESES}} & \multicolumn{5}{c||}{\textbf{French 2D UMAP}} & \multicolumn{5}{c|}{\textbf{English 2D UMAP}} \\ 
                 \cline{2-11}
 & \multicolumn{4}{c|}{\textbf{Local}} & \textbf{Global} &  \multicolumn{4}{c|}{\textbf{Local}} & \textbf{Global} \\
   \cline{2-11}
        & \multicolumn{2}{c|}{\textbf{Topic} (T)} & \multicolumn{2}{c|}{\textbf{Style} (S)} & \multirow{2}{*}{(T-S)} & \multicolumn{2}{c|}{\textbf{Topic} (T)} & \multicolumn{2}{c|}{\textbf{Style} (S)} & \multirow{2}{*}{(T-S)} \\ 

        \cline{2-5}\cline{7-10}
        \textit{Are the predictions checked?} & (T') & (T'') & (S') & (S'') & & (T') & (T'') & (S') & (S'') & \\ 
        \hline
        \texttt{mistral-embed} & \ding{51}$^{**}$ & \ding{51}$^{**}$ & \ding{55}$^{**}$ & \ding{51}$^{**}$ & \ding{51}$^{**}$ & \ding{51}$^{**}$ & \ding{51}$^{**}$ & \ding{55}$^{**}$ & \ding{51}$^{**}$ & \ding{51}$^{**}$ \\ 
        \hline
        \texttt{solon-...-large-0.1} & \ding{51}$^{**}$ & \ding{51}$^{**}$ & \ding{55} & \ding{51}$^{**}$ & \ding{51}$^{**}$ & \ding{51}$^{**}$ & \ding{51}$^{**}$ & \ding{55}$^{**}$ & \ding{51}$^{**}$ & \ding{51}$^{**}$ \\ 
        \hline
        \texttt{multilingual-e5-large} & \ding{51}$^{**}$ & \ding{51}$^{**}$ & \ding{51}$^{**}$ & \ding{51}$^{**}$ & \ding{51}$^{**}$ & \ding{51}$^{**}$ & \ding{51}$^{**}$ & \ding{55}$^{**}$ & \ding{51}$^{**}$ & \ding{51}$^{**}$ \\ 
        \hline
        \texttt{e5-base-v2} & \ding{51}$^{**}$ & \ding{51}$^{**}$ & \ding{51}$^{}$ & \ding{51}$^{**}$ & \ding{51}$^{**}$ & \ding{51}$^{**}$ & \ding{51}$^{**}$ & \ding{55} & \ding{51}$^{**}$ & \ding{51}$^{**}$ \\ 
        \hline
        \texttt{voyage-2} & \ding{51}$^{**}$ & \ding{51}$^{**}$ & \ding{55} & \ding{51}$^{**}$ & \ding{51}$^{**}$ & \ding{51}$^{**}$ & \ding{51}$^{**}$ & \ding{55}$^{**}$ & \ding{51}$^{**}$ & \ding{51}$^{**}$ \\ 
        \hline
        \texttt{sentence-camembert-base} & \ding{51}$^{**}$ & \ding{51}$^{**}$ & \ding{51}$^{**}$ & \ding{51}$^{**}$ & \ding{51}$^{**}$ & \ding{51}$^{**}$ & \ding{51}$^{**}$ & \ding{51}$^{**}$ & \ding{51}$^{**}$ & \ding{51} \\ 
        \hline
        \texttt{all-MiniLM-L12-v2} & \ding{51}$^{**}$ & \ding{51}$^{**}$ & \ding{51}$^{**}$ & \ding{51}$^{**}$ & \ding{51}$^{**}$ & \ding{51}$^{**}$ & \ding{51}$^{**}$ & \ding{51}$^{}$ & \ding{51}$^{**}$ & \ding{51}$^{**}$ \\ 
        \hline
        \texttt{text-embedding-3-small} & \ding{51}$^{**}$ & \ding{51}$^{**}$ & \ding{51}$^{**}$ & \ding{51}$^{**}$ & \ding{51}$^{**}$ & \ding{51}$^{**}$ & \ding{51}$^{**}$ & \ding{55}$^{*}$ & \ding{51}$^{**}$ & \ding{51}$^{**}$ \\ 
        \hline
        \texttt{-multi...-mpnet-base-v2} & \ding{51}$^{**}$ & \ding{51}$^{**}$ & \ding{55}$^{**}$ & \ding{51}$^{**}$ & \ding{51}$^{**}$ & \ding{51}$^{**}$ & \ding{51}$^{**}$ & \ding{51}$^{}$ & \ding{51}$^{**}$ & \ding{51}$^{**}$ \\ 
        \hline
        \texttt{xlm-roberta-large} & \ding{51}$^{**}$ & \ding{51}$^{**}$ & \ding{51}$^{**}$ & \ding{51}$^{**}$ & \ding{55}$^{**}$ & \ding{51}$^{**}$ & \ding{51}$^{**}$ & \ding{51}$^{**}$ & \ding{51}$^{**}$ & \ding{55}$^{**}$ \\ 
        \hline
        \texttt{all-roberta-large-v1} & \ding{55}$^{}$ & \ding{51}$^{**}$ & \ding{51}$^{**}$ & \ding{51}$^{**}$ & \ding{55}$^{**}$ & \ding{51}$^{**}$ & \ding{51}$^{**}$ & \ding{51}$^{**}$ & \ding{51}$^{**}$ & \ding{51}$^{**}$ \\ 
        \hline
        \texttt{distilbert-base-uncased} & \ding{51}$^{**}$ & \ding{51}$^{**}$ & \ding{51}$^{**}$ & \ding{51}$^{**}$ & \ding{55}$^{**}$ & \ding{51}$^{**}$ & \ding{55}$^{**}$ & \ding{55}$^{**}$ & \ding{51}$^{**}$ & \ding{51}$^{**}$ \\ 
        \hline
        \textbf{Mean} & \ding{51}$^{**}$ & \ding{51}$^{**}$ & \ding{51}$^{**}$ & \ding{51}$^{**}$ & \ding{51}$^{**}$ & \ding{51}$^{**}$ & \ding{51}$^{**}$ & \ding{55}$^{**}$ & \ding{51}$^{**}$ & \ding{51}$^{**}$ \\ 
        \hline
\end{tabular}}
\caption{Results of validation for hypotheses (T), (S) and (T-S) based on 2D UMAP projection for French and English. ``\ding{51}'' indicates that the prediction is verified, ``\ding{55}'' indicates that the opposite prediction is verified, with $^{*}$ and $^{**}$ reporting $p$-value $<.05$ and $<.01$, respectively.}
\label{tab:2dumap}
\end{table*}

For the $j$-th iteration, we  define $d_X^{(i,j)}$ as the Euclidean distance of the $i$-th embedding vector from the centroid $c_X^{(j)}$ of class $X$ as follows:

\begin{equation}\label{eq:euclid_iter}
d_X^{(i,j)} = \|v_X^{(i,j)} - c_X^{(j)} \|
\end{equation}

where $v_X^{(i,j)}$ is the $i$-th embedding vector of class $X$ in the $j$-th iteration, $c_X^{(j)}$ is the centroid vector for class $X$ in the $j$-th iteration, and $\| \cdot \|$ is the Euclidean norm.

To capture the spatial distribution of high-dimensional embeddings, we calculate the mean Euclidean distance from the centroid of each class across all iterations, written $\bar{d}_X(i)$:

\begin{equation}\label{eq:avg_distance}
\bar{d}_X(i) = \frac{1}{30} \sum_{j=1}^{30} d_X^{(i,j)}
\end{equation}

Finally, the overall mean distance $\bar{d}_X$ for class $X$ across all embeddings is:

\begin{equation}\label{eq:centroid}
\bar{d}_X = \frac{1}{N} \sum_{i=1}^{N} \bar{d}_X(i)
\end{equation}

where $\bar{d}_X(i)$ is the averaged Euclidean distance of the $i$-th embedding vector for class $X$ and $N$ is the total number of embedding vectors in the class.

In order to analyze the influence of topic variation, we compared the difference in embedding dispersion between classes presenting topic homogeneity versus topic heterogeneity, i.e. by comparing $\textsc{Queneau\_ref}$ with $\textsc{Feneon\_gen}$, and also $\textsc{Queneau\_gen}$ with $\textsc{Feneon\_ref}$. To analyze the influence of style, we compared the difference in embedding dispersion between classes showing style homogeneity versus style heterogeneity, i.e. by comparing $\textsc{Feneon\_ref}$ with $\textsc{Feneon\_gen}$, and also $\textsc{Queneau\_gen}$ with $\textsc{Queneau\_ref}$.

Using the metric defined in \eqref{eq:centroid}, we predict that both topic and writing style influence embeddings dispersion, as in the \textit{local hypotheses} (T) and (S):

\vspace{0.9pc}
\begin{subequations} \label{eq:5}$
    \makebox[0pt][l]{\small{Topic}}\quad\quad\left|
    \resizebox{0.73\columnwidth}{!}{ 
        $\begin{aligned}
            & \bar{d}_\textsc{Feneon\_gen} > \bar{d}_\textsc{Queneau\_ref} & \text{(T')}\\
            & \bar{d}_\textsc{Feneon\_ref} > \bar{d}_\textsc{Queneau\_gen} & \text{(T'')} \label{eq:5doubleprime}
        \end{aligned}$
    }\ \ {(\text{T})}\right.$
\end{subequations}

\vspace{0.9pc}
\begin{subequations}\label{eq:6}$
    \makebox[0pt][l]{\small{Style}}\quad\quad\left|
    \resizebox{0.73\columnwidth}{!}{ 
        $\begin{aligned}
            & \bar{d}_\textsc{Queneau\_ref} > \bar{d}_\textsc{Queneau\_gen} & \text{(S')} \label{eq:6prime}\\
            & \bar{d}_\textsc{Feneon\_gen} > \bar{d}_\textsc{Feneon\_ref} & \text{(S'')} \label{eq:6doubleprime}
        \end{aligned}$
    }\ \ {(\text{S})}\right.$
\end{subequations}
\vspace{0.5pc}

Besides hypotheses (T) and (S), we expect the topic to have a greater impact on embeddings dispersion than style, as defined in the \textit{global hypothesis} (T-S):

\vspace{0.9pc}
\begin{subequations}\label{eq:6}$
    \makebox[0pt][l]{\hspace{0.8cm}
    \resizebox{0.6\columnwidth}{!}{ 
        $\begin{aligned}
      \bar{d}_{_\textsc{Feneon\_ref}} > \bar{d}_{_\textsc{Queneau\_ref}} 
        \end{aligned}$
    }\hfill \hspace{0.9cm}{(\text{T-S})}}$
\end{subequations}
\vspace{0.9pc}

According to (T-S), the greater dispersion of $\textsc{Feneon\_ref}$ compared to $\textsc{Queneau\_ref}$ implies that topic variation influences more embedding dispersion than style variation, as the topic shifts to pluriform from $\textsc{Queneau\_ref}$ to $\textsc{Feneon\_ref}$, whereas style, in contrast, moves toward uniformity. 


Optimal hypotheses validation was obtained with 2D UMAP for both French and English, followed by weaker but still highly consistent results with 5D and 10D UMAP. To help visualize dispersion around centroids, Figure \ref{fig:dispersionplot} shows the contour plots of the 2D UMAP projected embeddings obtained with $\texttt{all-MiniLM-L12-v2}$ on the $\textsc{Queneau-Feneon}$ corpus. Detailed 2D UMAP results per hypothesis (T), (S) and (T-S) of the twelve models are given in Table \ref{tab:2dumap}.

For both English and French, mean centroid distances reported in Figure \ref{fig:dispersionplot} result in the following order: $\bar{d}_{_\textsc{Feneon\_gen}} > \bar{d}_{_\textsc{Feneon\_ref}} > \bar{d}_{_\textsc{Queneau\_ref}} > \bar{d}_{_\textsc{Queneau\_gen}}$. All pairwise comparisons of this order using two-sided t-tests proved to be significant at the .01 level. Locally, this validates (T')-(T''), supporting hypothesis (T) on the increasing effect of topic variation on embedding dispersion. Moreover, pairwise comparisons also significantly validate (S')-(S''), now verifying prediction (S) on the positive effect of writing style on vector dispersion. Globally, validation of (T-S) reveals a stronger effect of topic on this dispersion compared to style.

Looking at results individually given in Table \ref{tab:2dumap}, we observe that a vast majority of models validates hypotheses (T), (S) and (T-S) in both languages. Concerning results for French $\textsc{Queneau-Feneon}$, the dispersion effect predicted by (T')-(T'') for topic is consistently confirmed across models, with strong significance except in one case ($\texttt{all-roberta-large-v1}$). The dispersion effect of style predicted by (S')-(S'') was also significantly verified, except in 4 cases for condition (S'): $\texttt{mistral-embed}$, $\texttt{solon-embeddings-large-0.1}$, $\texttt{-multi-}$ $\texttt{lingual-mpnet-base-v2}$, and $\texttt{voyage-2}$. In support of the global hypothesis (T-S), varying the topic resulted in greater dispersion than varying the style for 9 (out of 12) models, with the 3 models showing the opposite direction.

The results obtained for English are largely similar to the results for French. Except in one single case ($\texttt{distilbert-base-uncased}$), the topic hypo\-theses (T) are significantly verified across all tested models. Concerning the style hypo\-the\-ses (S), results were more mixed for (S') with 6 models significantly invalidating the prediction but still strongly verified for (S'') across all models. 


In conclusion, the models applied on the French corpus outperformed their English counterparts in 2D UMAP dimension. The models that consistently satisfied both the topic and style hypotheses (T) and (S) in both languages were $\texttt{sentence-camembert-base}$, \mbox{$\texttt{all-MiniLM-L12-v2}$, and $\texttt{xlm-roberta-}$}
$\texttt{large}$. That said, the global hypothesis (T-S) is verified more on the English corpus than on the French corpus, with only one model rejecting the hypothesis in English ($\texttt{xlm-roberta-large}$).


\subsection{Style embedding interpretability}
\label{ssec:interpretability}

In the previous section, we observed that style variation, in addition to the more common topic variation, also influences embedding dispersion. Here we attempt to identify the key stylistic features related to style hypothesis (S), that may drive embedding vectors to exhibit greater or lesser dispersion.   

To conduct this analysis, we used a framework developed by \citet{terreau2021writing} to evaluate how embedding vectors represent writing style.\footnote{\url{https://github.com/EnzoFleur/style_embedding_evaluation/}} This framework generates stylometric reports to assess how well embedding models recognize writing styles in alignment with stylistic features identified by well-known Python modules (e.g. spaCy, NLTK, Counter). \citet{terreau2021writing} select eight groups of stylistic features as predictive targets for French and English regression models. These features include: the relative frequency of \textit{function words} (e.g., prepositions, conjunctions, auxiliary verbs) compared to the total word count in the text, the average values of \textit{structural features} (e.g., word length, word frequency, syllables per word), \textit{indexes} of lexical complexity (e.g., Yule’s K constancy measure \cite{yule2014statistical}, Shannon Entropy \cite{shannon1948mathematical}) and text readability metrics (e.g., Flesch-Kincaid Grade Level \cite{kincaid1975derivation}), the relative frequency of \textit{punctuation marks} (e.g., periods, commas) compared to total text length, \textit{numbers} (i.e., numerical digits), the average frequency of \textit{named entities} (i.e., NER: persons, locations, organizations) per sentence, and \textit{part-of-speech tags} (i.e., TAG: nouns, verbs, adjectives). 

Our main focus in this section is to analyze how embedding dispersion responds to variations in these stylistic features when style varies across classes, while the topic remains constant. In more detail, we focus on examining the interaction between differences in the frequencies of the eight stylistic features and the difference in dispersion between $\textsc{Queneau\_gen}$ and $\textsc{Queneau\_ref}$. This comparison is the only case where styles are expected to differ significantly between classes, while the topic remains exactly the same. As a control, we also compared $\textsc{Queneau\_gen}$ and $\textsc{Feneon\_ref}$, where the topics are expected to differ significantly, while the style remains the same.


We first measured mean ground frequencies of the eight stylistic features, written $\bar{f}^\text{s}$ with \textit{s} a stylistic feature, by applying~\citet{terreau2021writing}'s extraction module on the three classes of interest: $\textsc{Queneau\_gen}$, $\textsc{Queneau\_ref}$ and $\textsc{Feneon\_ref}$. For each of the two targeted comparisons, a pairwise t-test was conducted for each feature variation to assess statistical significance. Table \ref{tab:groundavg} presents the changes in feature frequencies from $\textsc{Queneau\_gen}$ to $\textsc{Queneau\_ref}$, and from $\textsc{Queneau\_gen}$ to $\textsc{Feneon\_ref}$.

\renewcommand{\arraystretch}{1.15}
\begin{table}[h]
\centering
\resizebox{1\columnwidth}{!}{%
\begin{tabular}{|c|l|c|c|c|c|}
\hline
& \multirow{2}{*}{\hspace*{2.5em}$\bar{f}^\text{s}$} & \multicolumn{2}{c|}{\textbf{to} $\textsc{Queneau\_ref}$} & \multicolumn{2}{c|}{\textbf{to} $\textsc{Feneon\_ref}$} \\
\cline{3-6}
\multirow{10}{*}{\rotatebox{90}{\textbf{From} $\textsc{Queneau\_gen}$}} &  & \multicolumn{1}{c|}{\textbf{French}} & \multicolumn{1}{c|}{\textbf{English}} & \multicolumn{1}{c|}{\textbf{French}} & \multicolumn{1}{c|}{\textbf{English}} \\
\cline{2-6}
&Function words   & $\nearrow^{**}$ & $\searrow^{**}$ & $\searrow^{**}$ & $\searrow^{*}$ \\
\cline{2-6}
&Indexes  & $\searrow_{**}$ & $\searrow_{**}$ & $\nearrow$ & $\nearrow$ \\
\cline{2-6}
&Letters  & $\nearrow$ & $\searrow_{**}$ & $\nearrow$ & $\searrow$ \\
\cline{2-6}
&NER  & $\nearrow^{}$ & $\nearrow^{*}$ & $\nearrow^{**}$ & $\nearrow^{**}$ \\
\cline{2-6}
&Numbers & $\nearrow$ & $\nearrow$ & $\nearrow^{**}$ & $\nearrow^{*}$ \\
\cline{2-6}
&Punctuation  & $\nearrow^{**}$  & $\searrow^{**}$ & $\nearrow$ & $\searrow$ \\
\cline{2-6}
&Structural & $\nearrow^{**}$ &  $\nearrow^{**}$ &  $\nearrow^{**}$ &  $\nearrow^{**}$ \\
\cline{2-6}
&TAG & $\nearrow^{}$ & $\nearrow^{*}$ & $\nearrow^{**}$ & $\nearrow^{**}$ \\

\hline
\end{tabular}}
\caption{Evolution of mean ground frequencies ($\bar{f}^\text{s}$) per text for the 8 stylistic features from $\textsc{Queneau\_gen}$ to $\textsc{Queneau\_ref}$ and $\textsc{Feneon\_ref}$, with $^{*}$ and $^{**}$ reporting $p$-value $<.05$ and $<.01$, respectively.}
\label{tab:groundavg}
\end{table}

For both French and English, we observed that structural features were significantly more frequent in $\textsc{Queneau\_ref}$ than in $\textsc{Queneau\_gen}$, and that indexes were significantly less frequent in $\textsc{Queneau\_ref}$ compared to $\textsc{Queneau\_gen}$. Opposite tendencies were observed for function words and punctuation between French (where both significantly increase) and English (where both significantly decrease). Additionally, some variations observed as non-significant in French (e.g. for letters, NER, TAG) were found to be significant in English. Moving to the comparison between $\textsc{Queneau\_gen}$ and $\textsc{Feneon\_ref}$, we observed that function words were significantly less frequent in $\textsc{Feneon\_ref}$ compared to $\textsc{Queneau\_gen}$, but that NER, numbers, structural and TAG became significantly more frequent in $\textsc{Feneon\_ref}$. The differences observed for indexes, letters and punctuation were not significant.


From now on, we restrict our attention only to differences \textit{observed as significant} for ground frequencies (see Table \ref{tab:groundavg}, cases marked with * or ** only). To see how these differences interact with differences in dispersion, we use the non-averaged $\Delta$ values per class, denoted as $\Delta d$ and $\Delta f^\text{s}$ respectively, and defined as follows: 

\begin{equation}
\Delta d (X,Y)= d_X(i) - d_Y(j)
\end{equation}
\begin{equation}
\Delta f^\text{s} (X,Y)= f^{s}_X(i) - f^{s}_Y(j)
\vspace{0.1cm}
\end{equation}

where $Y=\textsc{Queneau\_gen}$, $X$ is either $\textsc{Queneau\_ref}$ or $\textsc{Feneon\_ref}$ depending on the targeted comparison, $i$ is the $i$-th vector of class $X$, and $j$ is the $j$-th vector of class $Y$.

\begin{figure}[h]
    \centering
    \includegraphics[width=\linewidth]{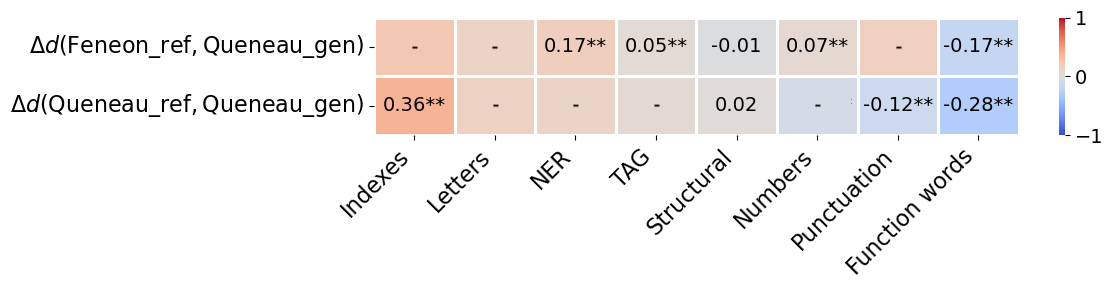}
    \label{fig:corr_plot_deltas_fr}
    
    \vspace{-0.3cm} 
    
    \includegraphics[width=\linewidth]{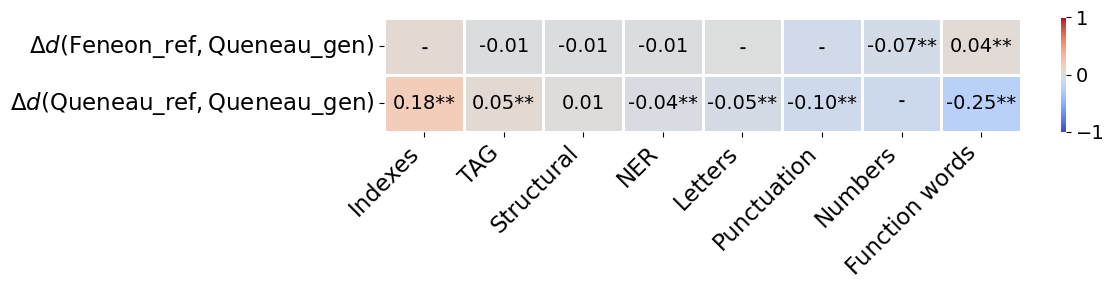}
    \label{fig:corr_plot_deltas_en}
      \vspace{-0.5cm} 
      
    \caption{Correlation matrices between differences in dispersion ($\Delta d$) and differences in frequencies of the eight stylistic features ($\Delta f^{s}$) for the two comparisons of interest, for French (top) and English (bottom). Here ``-'' corresponds to correlations that were intentionally omitted, as they correspond to differences in features previously observed as non significant (see Table \ref{tab:groundavg}).}
    \label{fig:corr_plot_deltas}
\end{figure}

Figure \ref{fig:corr_plot_deltas} reports the Pearson correlation coefficients measuring the interaction between the difference $\Delta d$ (i.e., the difference in embedding dispersion per text) and the difference $\Delta f^\text{s}$ (i.e., the difference in the frequency of stylistic features per text, for each stylistic feature $s$).

For $\textsc{Queneau\_gen}$ and $\textsc{Queneau\_ref}$ intended to differ in style, we observed moderate positive correlation in French between dispersion and differences in frequencies for indexes ($r=0.36^{**}$), and weak negative correlation with function words ($r=-0.28^{**}$) and punctuation ($r=-0.12^{**}$). No significant correlation was observed for structural ($r=0.02$). In English, for the same comparison, we observed only weak positive correlation with indexes ($r=0.18^{**}$), and weak negative correlation with function words ($r=-0.25^{**}$) and punctuation ($r=-0.10^{**}$). No correlation was observed in English with other features (e.g. TAG, NER, letters and structural).

Concerning $\textsc{Queneau\_gen}$ and $\textsc{Feneon\_ref}$ now intended \textit{not} to differ in style, we observed only weak positive correlation in French with NER ($r=0.17^{**}$), and weak negative correlation with function words ($r=-0.17^{**}$). No correlation was found with numbers ($r=0.07^{**}$), TAG ($r=0.05^{**}$) and structural ($r=-0.01$). In English, no correlation was observed at all.

To summarize, when corpora were expected \textit{not} to differ in writing style, such as $\textsc{Queneau\_gen}$ and $\textsc{Feneon\_ref}$, we observed weak or no correlation with dispersion for French, and no correlation at all for English. These findings align with expectations. By contrast, when corpora were expected to differ in style, such as $\textsc{Queneau\_gen}$ and $\textsc{Queneau\_ref}$, significant differences were observed for indexes, punctuation and function words. Also aligned with expectations, this result may account for the differences observed in dispersion between $\textsc{Queneau\_gen}$ and $\textsc{Queneau\_ref}$, indicating sensitivity of embedding vectors to these specific features in both French and English. 

While both languages exhibited sensitivity in the comparison between $\textsc{Queneau\_gen}$ and $\textsc{Queneau\_ref}$, the correlations were stronger for French than for English. Similarly, when comparing $\textsc{Queneau\_gen}$ and $\textsc{Feneon\_ref}$, the same variations in frequency were observed in both languages (i.e., function words, NER, numbers, structural and TAG), but language models were insensitive to these features in English, contrary to French. Finally, for variations significant in English only (i.e., letters, NER and TAG), no correlation with dispersion was observed. That said, Table \ref{tab:groundavg} offers a possible explanation for these discrepancies. Rather than reflecting an inherent limitation of language models with English, the directions of variations suggest that the translation from French to English has notably reduced the presence of function words, letters, and punctuation, while only slightly increasing the significance of features like NER and TAG. Accordingly, a plausible explanation for the reduced sensitivity in English is that translation may have diminished the first set of features to a degree that embedding vectors struggle to capture, without sufficiently amplifying the second set of features to balance this effect.

\section{Conclusion and perspectives}
\label{sec:conclusion}

This paper provides evidence that writing style influences embedding dispersion, though topic variation has a stronger effect. This result is supported across different language models in both French and English. Attempt at interpretability suggests that specific linguistic features, particularly readability and complexity indexes, function words and punctuation (to a lesser extent), partially explain embedding representations.

In the short run, two steps of investigation emerge. Firstly, 
some models we tested (e.g., $\texttt{sentence-camembert-base}$, $\texttt{all-Mini}$ $\texttt{LM-L12-v2}$) showed greater responsiveness to stylistic variations leading to increased dispersion. Other models (e.g., $\texttt{voyage-2}$, $\texttt{solon-embeddings-large-0.1}$) were less or not affected. This observation, including cases where dispersion decreased contrary to expectations (e.g., $\texttt{mistral-embed}$), suggests that different architectures process stylistic features in unique ways. Secondly, models generally responded more to stylistic variation in French compared to English. This was also reflected in the weaker stylistic correlations observed for English, suggesting that factors such as translation or language-specific characteristics could play a role. Replicating the study on a larger French-English corpus would provide further insight into these differences \textcolor{black}{and the detailed stylistic features controlling them}.

\textcolor{black}{In the long run, we aim for generalizability across other models and genres, also hoping to inspire further stylistic studies on languages that are more typologically diverse than French and English. Regarding models, we aim to compare the sensitivity of the architectures considered, with a focus on open-weight models to enhance explainability. Concerning genres, we aim to apply our methodology to news articles, as a genre responding to stylistic conventions other than literary conventions, associated to a great variety of topics and a potential for high scalability. We leave these investigations for future work.}

\section*{Limitations}
\label{sec:limitations}

Our corpus methodology is validated by clustering. However, the corpus is limited to 292 textual documents in each language (a total of 584 documents), with 73 documents per class. A larger dataset should be used for replication in order to mitigate biases due to sample size and verify our hypotheses further. 

While UMAP dimensionality reduction to 2D produced significant results in French, comparable results were also observed with higher UMAP dimensions. We omitted these in the paper for brevity but they are available on our GitHub repository (see supplementary materials below).

Additionally, our study is focused on eight main stylistic features, but these features are driven by subfeatures that should be considered to gain a clearer understanding of their impact on embedding dispersion. Other areas like journalism and scientific writing, which follow different stylistic conventions, are not explored either. Testing our hypotheses on other types of textual documents and assessing relevance of domain-specific features will be essential for assessing the generalizability of our results. 

Lastly, open-weight LLMs that we tested, like DistilBERT and RoBERTa, offer potential for deeper explainability and customization, in contrast to proprietary models, like OpenAI embeddings, that lack transparency due to their closed-weights. Large open-weight models like LLaMA-2 and Mistral-7B do exist, but their use requires significant computational resources.

\section*{Ethical considerations}
\label{sec:ethical}

Our research adheres to the following ethical principles: open science, transparency, inclusiveness, and sustainability. As academic researchers, we adhere to open science guidelines, with a concern for the reproducibility of experiments and the accessibility of our results. The dataset we provide contains no textual documents from the original sources, only vector embeddings derived from those documents, aligning with the concept of ``transformative fair use''. This approach ensures compliance with intellectual property and data protection regulations. Transparency is upheld through raw data availability, code and prompts sharing on a dedicated GitHub repository, and a comprehensive documentation to ensure reproducibility by others. We are also guided by inclusiveness, as we expect our research project to contribute to advancing educational and cultural AI literacy on the interplay between writing style and embedding representations. \textcolor{black}{To promote sustainability, the GitHub of the study actually supports net-zero carbon initiatives by others based on our framework. We also prioritize using smaller, open-source pre-trained language models alongside larger ones, to reach a balance between carbon footprint and resource consumption.}

\section*{Supplementary materials}

Large language models used in the experiments included the 1536-dimensional model text-embedding-3-small\footnote{\url{https://platform.openai.com/docs/guides/embeddings}} by OpenAI, and the 1024-dimensional models mistral-embed\footnote{\url{https://docs.mistral.ai/capabilities/embeddings/}} by Mistral, voyage-2\footnote{\url{https://docs.voyageai.com/docs/embeddings}}  by Voyage, and the RoBERTa-based models xlm-roberta-large,\footnote{\url{https://huggingface.co/FacebookAI/xlm-roberta-large}} all-roberta-large-v1,\footnote{\url{https://huggingface.co/sentence-transformers/all-roberta-large-v1}} and multilingual-e5-large.\footnote{\url{https://huggingface.co/intfloat/multilingual-e5-large}} Smaller models included the 768-dimensional models e5-base-v2,\footnote{\url{https://huggingface.co/intfloat/e5-base-v2}} distilbert-base-uncased,\footnote{\url{https://huggingface.co/distilbert/distilbert-base-uncased}} 
all-MiniLM-L12-v2,\footnote{\url{https://huggingface.co/sentence-transformers/all-MiniLM-L12-v2}} the SBERT model sentence-camembert-base\footnote{\url{https://huggingface.co/dangvantuan/sentence-camembert-base}} and the multilingual model paraphrase-multilingual-mpnet-base-v2.\footnote{\url{https://huggingface.co/sentence-transformers/paraphrase-multilingual-mpnet-base-v2}} We also included solon-embeddings-large-0.1\footnote{\url{https://huggingface.co/OrdalieTech/solon-embeddings-large-0.1}} (1024D) by Solon as one of the best performing French embedding models \textcolor{black}{(see MTEB leaderboard on HuggingFace,\footnote{\url{https://huggingface.co/spaces/mteb/leaderboard}} at time of the submission: September 16, 2024}). 

\medskip
\noindent All the code (including prompts for generating the extended corpus in French and English), data, and analytical results are available at the following URL link: \url{https://github.com/evangeliazve/topic_style_embeddings_dispersion/tree/main}

\section*{Acknowledgements}

We thank three anonymous reviewers for helpful comments and feedback. EZ acknowledges Infopro Digital for supporting her PhD research, alongside her work. BI acknowledges the program THEMIS (grant agreements n°DOS022279400 and n°DOS022279500) for funding.

\section*{Declaration of contribution}
BI and JGG conceptualized the research problem and designed the experiment with EZ. LS and AB managed the data collection and generation processes. EZ was responsible of coding, optimizing, integrating existing frameworks and testing selected models. EZ  and BI analyzed and discussed the results. BI and EZ wrote the paper, which all authors read and revised together. BI and EZ share first authorship. Correspondence: benjamin.icard@lip6.fr, evangelia.zve@lip6.fr, jean-gabriel.ganascia@lip6.fr.

\bibliography{biblio}

\end{document}